\documentclass[10pt,twocolumn,letterpaper]{article}

\usepackage{3dv}
\usepackage{times}
\usepackage{epsfig}
\usepackage{graphicx}
\usepackage{amsmath}
\usepackage{amssymb}
\usepackage{bm}
\usepackage{lipsum}
\usepackage{balance}

\usepackage{enumitem}
\setlist[enumerate]{noitemsep, topsep=3pt}

\usepackage{placeins} % For FloatBarrier
\usepackage{flafter}  % For FloatBarrier

\usepackage{tabularx, booktabs}
\usepackage{multirow}
\newcolumntype{Y}{>{\centering\arraybackslash}X}

\usepackage[linesnumbered,ruled,vlined]{algorithm2e}

\SetCommentSty{mycommfont}
\SetKwInput{KwInput}{Input}                
\SetKwInput{KwOutput}{Output}

\usepackage[pagebackref=true,breaklinks=true,letterpaper=true,colorlinks,bookmarks=false]{hyperref}

\threedvfinalcopy % *** Uncomment this line for the final submission

\def\threedvPaperID{252} % *** Enter the 3DV Paper ID here

\begin{document}

%%%%%%%%% TITLE
\title{Geometric Interpretations of the Normalized Epipolar Error}

\author{Seong Hun Lee \hspace{25pt} Javier Civera \\
I3A, University of Zaragoza, Spain\\
{\tt\small \{seonghunlee, jcivera\}@unizar.es}
}

\maketitle
% \thispagestyle{empty}

%%%%%%%%% ABSTRACT
\begin{abstract}
\vspace{-0.5em}
    In this work, we provide geometric interpretations of the normalized epipolar error.
    Most notably, we show that it is directly related to the following quantities: 
    (1) the shortest distance between the two backprojected rays, 
    (2) the dihedral angle between the two bounding epipolar planes,
    and (3) the $L_1$-optimal angular reprojection error.
\end{abstract}

%%%%%%%%% BODY TEXT
\section{Introduction}
Consider two cameras $c_0$ and $c_1$ observing the same 3D point $\mathbf{p}$.
If the internal calibration and the relative pose of the two cameras are known, we can backproject the measured point in each image and obtain the rays from each camera, pointing to $\mathbf{p}$.
Now, we define the \textit{normalized epipolar error} as follows:
\vspace{-0.3em}
\begin{equation}
\label{eq:normalized_epipolar_error}
    \widehat{e} := \big|\hspace{1pt} \widehat{\mathbf{f}}_1\cdot\big(\hspace{2pt}\widehat{\mathbf{t}}\times\mathbf{R}\widehat{\mathbf{f}}_0\big)\big|
    =\big|\hspace{1pt} \widehat{\mathbf{t}}\cdot\big(\mathbf{R}\widehat{\mathbf{f}}_0\times\widehat{\mathbf{f}}_1\big)\big|,
    \vspace{-0.3em}
\end{equation}
where $\widehat{\mathbf{f}}_0$ and $\widehat{\mathbf{f}}_1$ are the backprojected unit rays from $c_0$ and $c_1$, respectively, $\mathbf{R}$ is the rotation matrix and $\mathbf{t}$ is the translation vector that together transform a point from the reference frame $c_0$ to $c_1$, i.e., $\mathbf{x}_1=\mathbf{Rx}_0+\mathbf{t}$ and $\widehat{\mathbf{t}}=\mathbf{t}/\lVert\mathbf{t}\rVert$.
The second equality in \eqref{eq:normalized_epipolar_error} follows from the fact that the scalar triple product is invariant to a circular shift.
In the literature, the error $\widehat{e}$ is often expressed as follows:
\vspace{-0.3em}
\begin{equation}
    \widehat{e} = \big|\hspace{1pt}\widehat{\mathbf{f}}_1^{\hspace{1.5pt}\top}\mathbf{E}\widehat{\mathbf{f}}_0\hspace{1pt}\big|,
    \vspace{-0.3em}
\end{equation}
where $\mathbf{E}=[\hspace{2pt} \widehat{\mathbf{t}}\hspace{2pt} ]_\times\mathbf{R}$ is the \textit{essential matrix} and $[\cdot]_\times$ is the skew-symmetric operator.

\begin{figure}[t]
 \centering
 \includegraphics[width=0.3\textwidth]{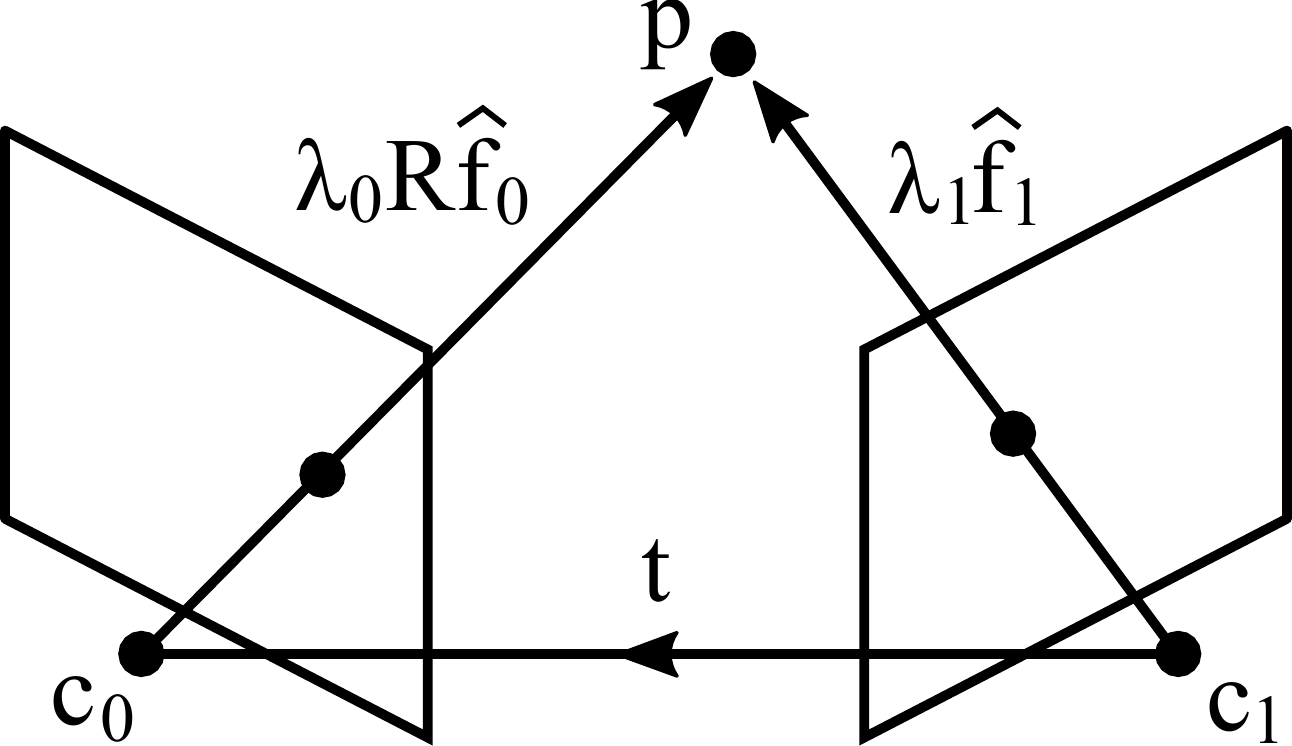}
\caption{A perfectly accurate epipolar geometry.
The two backprojected rays intersect at the exact position of the 3D point.
The depths of each ray are denoted by $\lambda_0$ and $\lambda_1$.
All vectors are expressed in the coordinate system of $c_1$.
 }
\label{fig:epipolar_perfect}
\end{figure}

If the image measurements, calibration and pose data are all perfectly accurate, this error would be zero because $\mathbf{R}\widehat{\mathbf{f}}_0$, $\widehat{\mathbf{f}}_1$ and $\widehat{\mathbf{t}}$ would be coplanar (see Fig. \ref{fig:epipolar_perfect}).
This is called the \textit{epipolar constraint} \cite{eight_point}.
In practice, the raw data contain inaccuracies, so they do not satisfy this constraint most of the time. 
% For this reason, many existing works in 3D vision try to solve geometric reconstruction problems by minimizing the cost based on this error \cite{certifiable, garcia2020certifiable, hlmke2007essential, kneip_direct_rotation, pagani2011structure, rodriguez2011reduced, zhao2019efficient}.
For this reason, many existing works in 3D vision try to solve geometric reconstruction problems by minimizing the cost based on this error \cite{certifiable, garcia2020certifiable, kneip_direct_rotation, roba, pagani2011structure, rodriguez2011reduced, zhao2019efficient, spetsakis1992optimal} or use this error to identify outliers \cite{zhao2019efficient, robust_uncertainty_aware_multiview_triangulation}. 

In the literature, the normalized epipolar error has mostly been treated as an algebraic quantity that has no geometric meaning \cite{certifiable, garcia2020certifiable, kneip_direct_rotation, rodriguez2011reduced,  zhao2019efficient, yang2014optimal}.
We believe that this misconception stems from the fact that the ``standard" epipolar error $e$ is an algebraic quantity \cite{hartley_book, luong1996fundamental,torr1997development,zhang1998determining}:
\vspace{-0.3em}
\begin{equation}
    \label{eq:epipolar_error}
    e := 
    \big| \mathbf{f}_1\cdot\big(\hspace{2pt}{\widehat{\mathbf{t}}}\times\mathbf{R}\mathbf{f}_0\big)\big|
    =
    \big|\mathbf{f}_1^\top\mathbf{E}\mathbf{f}_0\big|,
    \vspace{-0.3em}
\end{equation}
where $\mathbf{f}_0$ and $\mathbf{f}_1$ are the normalized image coordinates of the point in $c_0$ and $c_1$, respectively.
Notice that the only difference between \eqref{eq:normalized_epipolar_error} and \eqref{eq:epipolar_error} is the way the rays are normalized: 
In \eqref{eq:normalized_epipolar_error}, they are normalized by their lengths, whereas in \eqref{eq:epipolar_error}, they are normalized by the last element in the vector.

In \cite{pagani2011structure}, a geometric interpretation was given for the following error:
\vspace{-0.5em}
\begin{equation}
    e_p:= \frac{\big|\hspace{1pt}\widehat{\mathbf{f}}_1\cdot\big(\hspace{2pt}\widehat{\mathbf{t}}\times\mathbf{R}\widehat{\mathbf{f}}_0\big)\big|}{\lVert\hspace{2pt}\widehat{\mathbf{t}}\times\mathbf{R}\widehat{\mathbf{f}}_0\rVert},
    \vspace{-0.3em}
\end{equation}
which corresponds to the cosine of the angle between $\widehat{\mathbf{f}}_1$ and $\mathbf{n}=\widehat{\mathbf{t}}\times\mathbf{R}\widehat{\mathbf{f}}_0$.
This is equal to the perpendicular distance between the point at $\widehat{\mathbf{f}}_1$ and the plane containing $\widehat{\mathbf{t}}$ and $\mathbf{R}\widehat{\mathbf{f}}_0$.

In this work, we provide geometrically intuitive interpretations of \eqref{eq:normalized_epipolar_error} by relating it to the following quantities:
\begin{enumerate}\itemsep0em
    \item The volume of the tetrahedron where $\widehat{\mathbf{f}}_0$, $\mathbf{R}\widehat{\mathbf{f}}_0$ and $\widehat{\mathbf{t}}$ form the three edges meeting at one vertex.
    \item The shortest distance between the two backprojected rays $\mathbf{l}_0=\mathbf{t}+s_0\mathbf{R}\widehat{\mathbf{f}}_0$ and $\mathbf{l}_1=s_1\widehat{\mathbf{f}}_1$.
    \item The dihedral angle between the two bounding epipolar planes, \ie, one plane containing $\mathbf{t}$ and $\mathbf{R}\mathbf{f}_0$ and the other containing $\mathbf{t}$ and $\mathbf{f}_1$.
    \item The $L_1$-optimal angular reprojection error.
\end{enumerate}

\begin{figure}[t]
 \centering
 \includegraphics[width=0.475\textwidth]{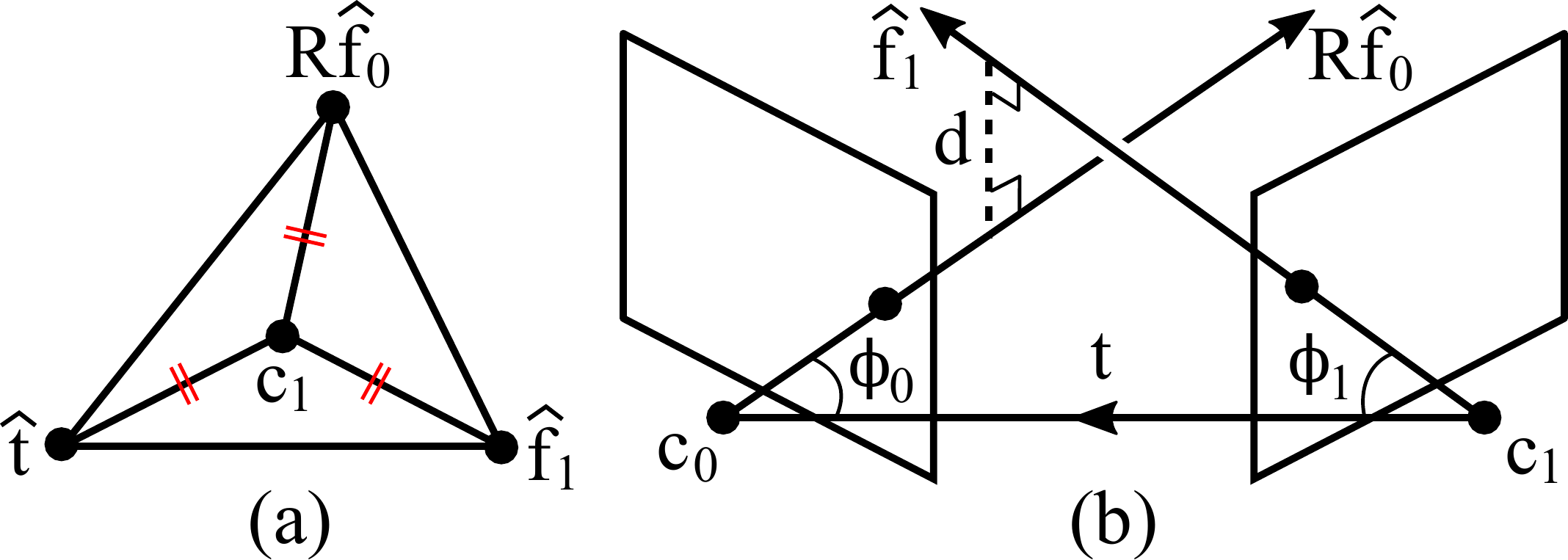}
\caption{\textbf{(a)} The volume of this tetrahedron is proportional to the normalized epipolar error \eqref{eq:normalized_epipolar_error}.
\textbf{(b)} $d$ is the shortest distance between the two backprojected rays corresponding to the same point.}
\label{fig:tetrahedron}
\end{figure}

\section{Geometric Interpretations of \eqref{eq:normalized_epipolar_error}}

\subsection{Relation to the volume of a tetrahedron}

Consider the tetrahedron shown in Fig. \ref{fig:tetrahedron}{\color{red}a}.
One of its vertices is placed at $\mathbf{c}_1$ (\ie, the position of camera $c_1$, which is the origin in the reference frame of $c_1$), and the other three at $\widehat{\mathbf{t}}$, $\mathbf{R}\widehat{\mathbf{f}}_0$ and $\widehat{\mathbf{f}}_1$.
Then, using the well-known formula for the volume of a tetrahedron, its volume is obtained by
\vspace{-0.2em}
\begin{equation}
    V = \frac{1}{6}\big|\hspace{1pt} \widehat{\mathbf{t}}\cdot\big(\hspace{2pt}\mathbf{R}\widehat{\mathbf{f}}_0\times\widehat{\mathbf{f}}_1\big)\big|\stackrel{\eqref{eq:normalized_epipolar_error}}{=}\frac{\widehat{e}}{6}.
    \vspace{-0.2em}
\end{equation}
Therefore, 
\vspace{-0.2em}
\begin{equation}
\label{eq:volume_interpretation}
    \widehat{e} = 6V.
    \vspace{-0.2em}
\end{equation}
The nice thing about this interpretation is that it allows for a simple visualization of the error, as shown in Fig. \ref{fig:tetrahedron}{\color{red}a}.
As the degree of coplanarity increases among the three edges ($\widehat{\mathbf{t}}$, $\mathbf{R}\widehat{\mathbf{f}}_0$ and $\widehat{\mathbf{f}}_1$), the common vertex will be ``pulled" towards the opposite side, flattening the tetrahedron.
When the three edges are coplanar, the tetrahedron becomes completely flat, \ie, $V=0$, and thus $\widehat{e}=0$.

\subsection{Relation to the distance between the two rays}

\begin{figure}[t]
 \centering
 \includegraphics[width=0.475\textwidth]{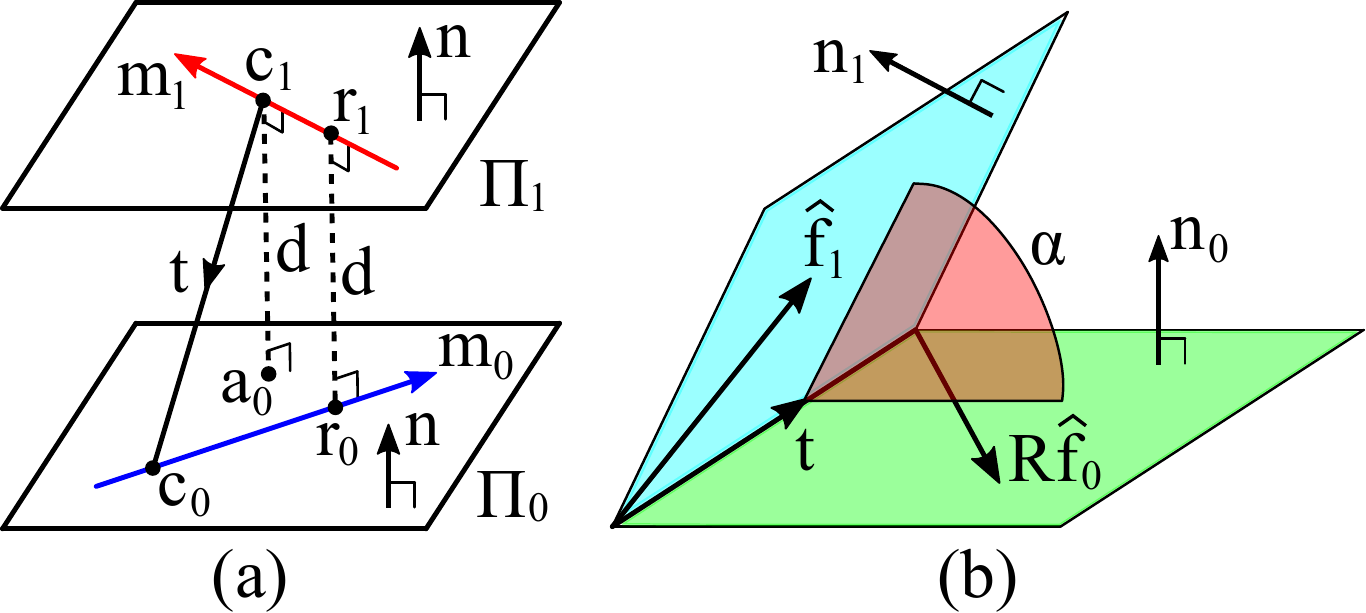}
\caption{\textbf{(a)} The distance between the two skew lines is the same as the distance between the two parallel planes as shown here.
The normal of the two planes is given by $\mathbf{n}=\mathbf{m}_0\times\mathbf{m}_1$.
\textbf{(b)} $\alpha$ is the dihedral angle between the two bounding epipolar planes.
 }
\label{fig:skew_distance}
\end{figure}

We can also relate the normalized epipolar error $\widehat{e}$ \eqref{eq:normalized_epipolar_error} to the shortest distance between the two backprojected rays, \ie, $d$ in Fig. \ref{fig:tetrahedron}{\color{red}b}.
To show this, we will first derive the formula for the shortest distance between two skew lines.
Consider two skew lines $\mathbf{l}_0=\mathbf{c}_0+s_0\mathbf{m}_0$ and $\mathbf{l}_1=\mathbf{c}_1+s_1\mathbf{m}_1$.
The distance between them is given by the distance between the closest pair of points on each line ($\mathbf{r}_0$ and $\mathbf{r}_1$), and they lie on the common perpendicular to both lines\footnote{We can easily prove this by contradiction. We omit the proof.}.
Now, consider two parallel planes with the normal $\mathbf{n}=\mathbf{m}_0\times\mathbf{m}_1$: plane $\Pi_0$ containing $\mathbf{l}_0$ and plane $\Pi_1$ containing $\mathbf{l}_1$, as illustrated in Fig. \ref{fig:skew_distance}{\color{red}a}.
Notice that $d=\lVert\mathbf{r}_0-\mathbf{r}_1\rVert$ is the same as the distance between the planes, which is the same as $\lVert\mathbf{c}_1-\mathbf{a}_0\rVert$ where $\mathbf{a}_0$ is the projected position of $\mathbf{c}_1$ in $\Pi_0$.
Since $\lVert\mathbf{c}_1-\mathbf{a}_0\rVert=|(\mathbf{c}_0-\mathbf{c}_1)\cdot\mathbf{n}|/\lVert\mathbf{n}\rVert$, we get
\vspace{-0.2em}
\begin{equation}
    d=\frac{|(\mathbf{c}_0-\mathbf{c}_1)\cdot(\mathbf{m}_0\times\mathbf{m}_1)|}{\lVert\mathbf{m}_0\times\mathbf{m}_1\rVert}.
    \vspace{-0.2em}
\end{equation}
This means that $d$ in Fig. \ref{fig:tetrahedron}{\color{red}b} is given by
\vspace{-0.3em}
\begin{equation}
\label{eq:distance}
    d=\frac{\big|\mathbf{t}\cdot(\mathbf{R}\widehat{\mathbf{f}}_0\times\widehat{\mathbf{f}}_1)\big|}{\lVert\mathbf{R}\widehat{\mathbf{f}}_0\times\widehat{\mathbf{f}}_1\rVert}
    \stackrel{\eqref{eq:normalized_epipolar_error}}{=}\frac{\lVert\mathbf{t}\rVert \hspace{1pt} \widehat{e}}{\lVert\mathbf{R}\widehat{\mathbf{f}}_0\times\widehat{\mathbf{f}}_1\rVert}.
    \vspace{-0.3em}
\end{equation}
Let $\beta$ be the angle between $\mathbf{R}\widehat{\mathbf{f}}_0$ and $\widehat{\mathbf{f}}_1$ (also known as the raw parallax angle \cite{triangulation_why_optimize}), \ie,
\vspace{-0.3em}
\begin{equation}
    \beta:=\angle(\mathbf{R}\widehat{\mathbf{f}}_0, \widehat{\mathbf{f}}_1)\in[0,\pi/2].
    \vspace{-0.3em}
\end{equation}
Then, \eqref{eq:distance} can be written as
\vspace{-0.3em}
\begin{equation}
\label{eq:distance_interpretation}
    \widehat{e}=\frac{\sin(\beta)}{\lVert\mathbf{t}\rVert}d.
    \vspace{-0.3em}
\end{equation}
Therefore, we can interpret $\widehat{e}$ as the distance between the two backprojected rays, weighted by $\sin(\beta)/\lVert\mathbf{t}\rVert$.
For relative pose estimation between two views, we can assume $\lVert\mathbf{t}\rVert=1$ without loss of generality, so minimizing the cost based on \eqref{eq:distance_interpretation} is equivalent to minimizing the cost based on $\sin(\beta)d$.
We can interpret $\sin(\beta)$ as the factor that downweights the residual $d$ when the parallax angle is small.
Note that $d\leq1$ and the equality holds if and only if $\mathbf{R}\widehat{\mathbf{f}}_0$ and $\widehat{\mathbf{f}}_1$ are both perpendicular to $\widehat{\mathbf{t}}$.
If the two rays intersect (at infinity), then $d=0$ (or $\beta=0$), and thus $\widehat{e}=0$.

As a side note, it should be mentioned that $d$ given by \eqref{eq:distance} is the distance between the \textit{lines} rather than the \textit{rays}.
Technically speaking, it is the shortest distance between line $\mathbf{l}_0=\mathbf{t}+s_0\mathbf{R}\widehat{\mathbf{f}}_0$ for $s_0\in\mathbb{R}$ and line $\mathbf{l}_1=s_1\widehat{\mathbf{f}}_1$ for $s_1\in\mathbb{R}$. 

\subsection{Relation to the angle between the two planes}

In Fig. \ref{fig:tetrahedron}{\color{red}b}, consider the following planes:
one plane containing $\mathbf{t}$ and $\mathbf{R}\widehat{\mathbf{f}}_0$, and another containing $\mathbf{t}$ and $\widehat{\mathbf{f}}_1$.
Let $\mathbf{n}_0$ and $\mathbf{n}_1$ be their normal vectors.
These two planes are drawn in Fig. \ref{fig:skew_distance}{\color{red}b}.
We can think of them as the two bounding planes between which the epipolar plane is usually found. 
This is the case for most two-view triangulation methods (\eg, midpoint methods \cite{triangulation_why_optimize} and optimal methods \cite{closed_form_optimal_triangulation_based_angular_errors}).
The dihedral angle between these two bounding planes is given by
\vspace{-0.3em}
\begin{equation}
    \alpha=\angle(\mathbf{n}_0, \mathbf{n}_1)=\sin^{-1}\left(\left\Vert\frac{\big(\hspace{1pt}\mathbf{R}\widehat{\mathbf{f}}_0\times\widehat{\mathbf{t}}\hspace{2pt}\big)}{\big\lVert\mathbf{R}\widehat{\mathbf{f}}_0\times\widehat{\mathbf{t}}\big\rVert}\times\frac{\big(\hspace{1pt}\widehat{\mathbf{f}}_1\times\widehat{\mathbf{t}}\hspace{2pt}\big)}{\big\lVert\widehat{\mathbf{f}}_1\times\widehat{\mathbf{t}}\big\rVert}\right\Vert\right).
    \vspace{-0.3em}
\end{equation}
This can be rearranged to
\begin{equation}
\label{eq:dihedral1}
\big\Vert\mathbf{R}\widehat{\mathbf{f}}_0\times\widehat{\mathbf{t}}\big\Vert  
\hspace{1pt}\big\Vert\widehat{\mathbf{f}}_1\times\widehat{\mathbf{t}}\big\Vert
\sin(\alpha) =
    \big\Vert\big(\hspace{1pt}\mathbf{R}\widehat{\mathbf{f}}_0\times\widehat{\mathbf{t}}\hspace{2pt}\big)\times\big(\hspace{1pt}\widehat{\mathbf{f}}_1\times\widehat{\mathbf{t}}\hspace{2pt}\big)\big\Vert.
\end{equation}
For any 3D vector $\mathbf{a}$, $\mathbf{b}$, $\mathbf{c}$, $\mathbf{d}$, the vector quadruple product $(\mathbf{a}\times\mathbf{b})\times(\mathbf{c}\times\mathbf{d})$ is equal to $\left((\mathbf{a}\times\mathbf{b})\cdot\mathbf{d}\right)\mathbf{c}-\left((\mathbf{a}\times\mathbf{b})\cdot\mathbf{c}\right)\mathbf{d}$.
Therefore, the right-hand side of \eqref{eq:dihedral1} can be written as
\begin{align}
    &\Big\Vert\big(\hspace{1pt}\mathbf{R}\widehat{\mathbf{f}}_0\times\widehat{\mathbf{t}}\hspace{2pt}\big)\times\big(\hspace{1pt}\widehat{\mathbf{f}}_1\times\widehat{\mathbf{t}}\hspace{2pt}\big)\Big\Vert \nonumber\\
    &\hspace{0.5em}=
    \Big\Vert\underbrace{\left((\mathbf{R}\widehat{\mathbf{f}}_0\times\widehat{\mathbf{t}})\cdot\widehat{\mathbf{t}}\right)}_{0}\widehat{\mathbf{f}}_1
    -
    \left((\mathbf{R}\widehat{\mathbf{f}}_0\times\widehat{\mathbf{t}}\hspace{2pt})\cdot\widehat{\mathbf{f}}_1\right)\widehat{\mathbf{t}}\hspace{1pt}\Big\Vert\\
    &\hspace{0.5em}=\left\Vert\left((\mathbf{R}\widehat{\mathbf{f}}_0\times\widehat{\mathbf{t}}\hspace{2pt})\cdot\widehat{\mathbf{f}}_1\right)\widehat{\mathbf{t}}\hspace{1pt}\right\Vert
    =\big|(\mathbf{R}\widehat{\mathbf{f}}_0\times\widehat{\mathbf{t}}\hspace{2pt})\cdot\widehat{\mathbf{f}}_1\big|
    \stackrel{\eqref{eq:normalized_epipolar_error}}{=} \widehat{e}. \label{eq:dihedral2}
\end{align}
Note that the third equality follows from the fact that $\widehat{\mathbf{t}}$ is a unit vector.
Substituting \eqref{eq:dihedral2} into \eqref{eq:dihedral1} leads to
\begin{align}
    \widehat{e}
    &=\big\lVert\mathbf{R}\widehat{\mathbf{f}}_0\times\widehat{\mathbf{t}}\big\rVert  
\big\lVert\widehat{\mathbf{f}}_1\times\widehat{\mathbf{t}}\big\rVert\sin(\alpha)\\
    &=\sin(\phi_0)\sin(\phi_1)\sin(\alpha), \label{eq:dihedral3}
\end{align}
where 
\begin{align}
    \phi_0&:=\angle(\mathbf{R}\widehat{\mathbf{f}}_0, \widehat{\mathbf{t}}\hspace{2pt})\in[0,\pi/2], \label{eq:phi0}\\
    \phi_1&:=\angle(\hspace{1pt}\widehat{\mathbf{f}}_1, \widehat{\mathbf{t}}\hspace{2pt})\in[0,\pi/2].\label{eq:phi1}
\end{align}
These two angles are shown in Fig. \ref{fig:tetrahedron}{\color{red}b}.
From \eqref{eq:dihedral3}, we can interpret $\widehat{e}$ as the sine of the dihedral angle between the two bounding epipolar planes, weighted by $\sin(\phi_0)\sin(\phi_1)$.
Therefore, $\widehat{e}$ would be small if either of $\phi_0$, $\phi_1$ or $\alpha$ is very small.
This makes sense because the epipolar geometry degenerates as $\phi_0$ or $\phi_1$ approaches zero.
Also, when $\alpha$ is small, the two bounding epipolar planes are close to coplanarity, and so do the vector $\widehat{\mathbf{t}}$, $\mathbf{R}\widehat{\mathbf{f}}_0$ and $\widehat{\mathbf{f}}_1$. 

\subsection{Relation to the angular reprojection error}

\begin{figure}[t]
 \centering
 \includegraphics[width=0.3\textwidth]{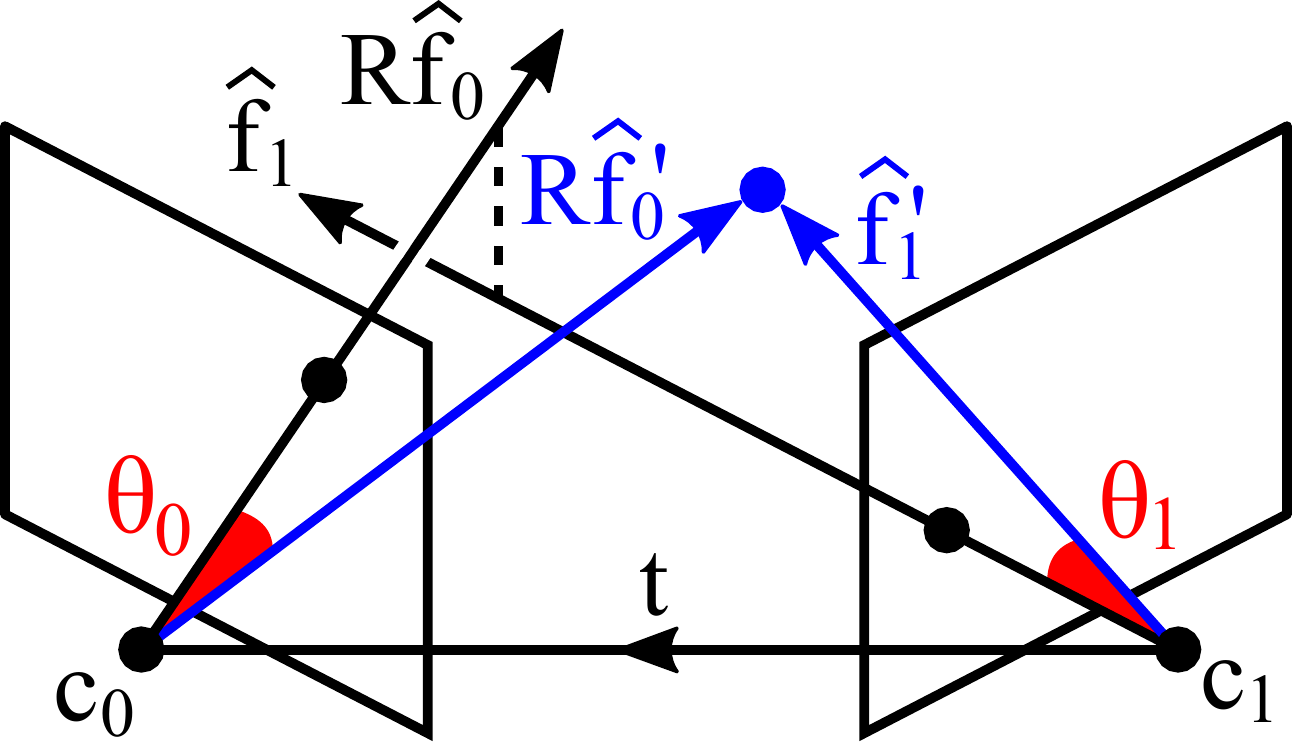}
\caption{The angular reprojection errors ($\theta_0$, $\theta_1$) measure the angular difference between the backprojected rays ($\mathbf{R}\widehat{\mathbf{f}}_0$, $\widehat{\mathbf{f}}_1$) and the corrected rays ($\mathbf{R}\widehat{\mathbf{f}}'_0$, $\widehat{\mathbf{f}}'_1$) that are made to intersect.
 }
\label{fig:angular}
\end{figure}

The $L_1$-optimal angular reprojection error \cite{closed_form_optimal_triangulation_based_angular_errors} is defined as follows:
\begin{equation}
\label{eq:L1_angle_definition}
    \theta^*_\mathrm{L1}=\min_{\widehat{\mathbf{f}}'_0, \widehat{\mathbf{f}}'_1}\left(\angle(\widehat{\mathbf{f}}'_0, \widehat{\mathbf{f}}_0)+\angle(\widehat{\mathbf{f}}'_1, \widehat{\mathbf{f}}_1)\right) \ \ \text{s.t.} \ \  \widehat{\mathbf{f}}'_1\cdot\big(\hspace{2pt}\widehat{\mathbf{t}}\times\mathbf{R}\widehat{\mathbf{f}}'_0\big)=0.
\end{equation}
In other words, it is the minimum of $\theta_0+\theta_1$ where $\theta_0$ and $\theta_1$ are the angles by which we correct the backprojected rays to make them intersect.
Fig. \ref{fig:angular} illustrates these angles.
In \cite{closed_form_optimal_triangulation_based_angular_errors}, it was shown that 
\begin{equation}
    \sin(\theta^*_\mathrm{L1})=\min\left(\frac{\big|\mathbf{R}\widehat{\mathbf{f}}_0\cdot\big(\hspace{2pt}\widehat{\mathbf{f}}_1\times\widehat{\mathbf{t}}\hspace{2pt}\big)\big|}{\lVert\mathbf{R}\widehat{\mathbf{f}}_0\times\widehat{\mathbf{t}}\rVert}, \frac{\big|\mathbf{R}\widehat{\mathbf{f}}_0\cdot\big(\hspace{2pt}\widehat{\mathbf{f}}_1\times\widehat{\mathbf{t}}\hspace{2pt}\big)\big|}{\lVert\widehat{\mathbf{f}}_1\times\widehat{\mathbf{t}}\rVert}\right).
\end{equation}
Rearranging this, we get
\begin{equation}
    \big|\hspace{1pt} \widehat{\mathbf{f}}_1\cdot\big(\hspace{2pt}\widehat{\mathbf{t}}\times\mathbf{R}\widehat{\mathbf{f}}_0\big)\big|=
    \sin(\theta^*_\mathrm{L1})\max\left(\lVert\mathbf{R}\widehat{\mathbf{f}}_0\times\widehat{\mathbf{t}}\rVert, \lVert\widehat{\mathbf{f}}_1\times\widehat{\mathbf{t}}\rVert\right).
\end{equation}
Using \eqref{eq:normalized_epipolar_error}, \eqref{eq:phi0}, \eqref{eq:phi1}, this can be written as
\begin{equation}
\label{eq:L1_angle_interpretation}
    \widehat{e}=\sin\left(\max(\phi_0,\phi_1)\right)\sin(\theta^*_\mathrm{L1}).
\end{equation}
Therefore, we can interpret $\widehat{e}$ as the sine of the $L_1$-optimal angular reprojection error, weighted by $\sin\left(\max(\phi_0,\phi_1)\right)$.
It follows that $\widehat{e}$ would be small if either of $\theta^*_\mathrm{L1}$ or $\max(\phi_0, \phi_1)$ is very small. 
This makes sense because small $\theta^*_\mathrm{L1}$ means that only a little correction is needed for the two backprojected rays to intersect. 
Also, small $\max(\phi_0, \phi_1)$ means that the vector $\widehat{\mathbf{t}}$, $\mathbf{R}\widehat{\mathbf{f}}_0$ and $\widehat{\mathbf{f}}_1$ are all close to parallelism, which brings the epipolar geometry close to degeneracy.
What may seem peculiar in \eqref{eq:L1_angle_interpretation} is the fact that $\max(\phi_0, \phi_1)$ does not reflect the degeneracy when either $\phi_0$ or $\phi_1$ is zero.
However, this is not an issue, because the term $\sin(\theta^*_\mathrm{L1})$ is necessarily zero whenever degeneracy occurs.
In the Appendix, we verify \eqref{eq:L1_angle_interpretation} using simulation.

\begin{figure}[t]
 \centering
 \includegraphics[width=0.44\textwidth]{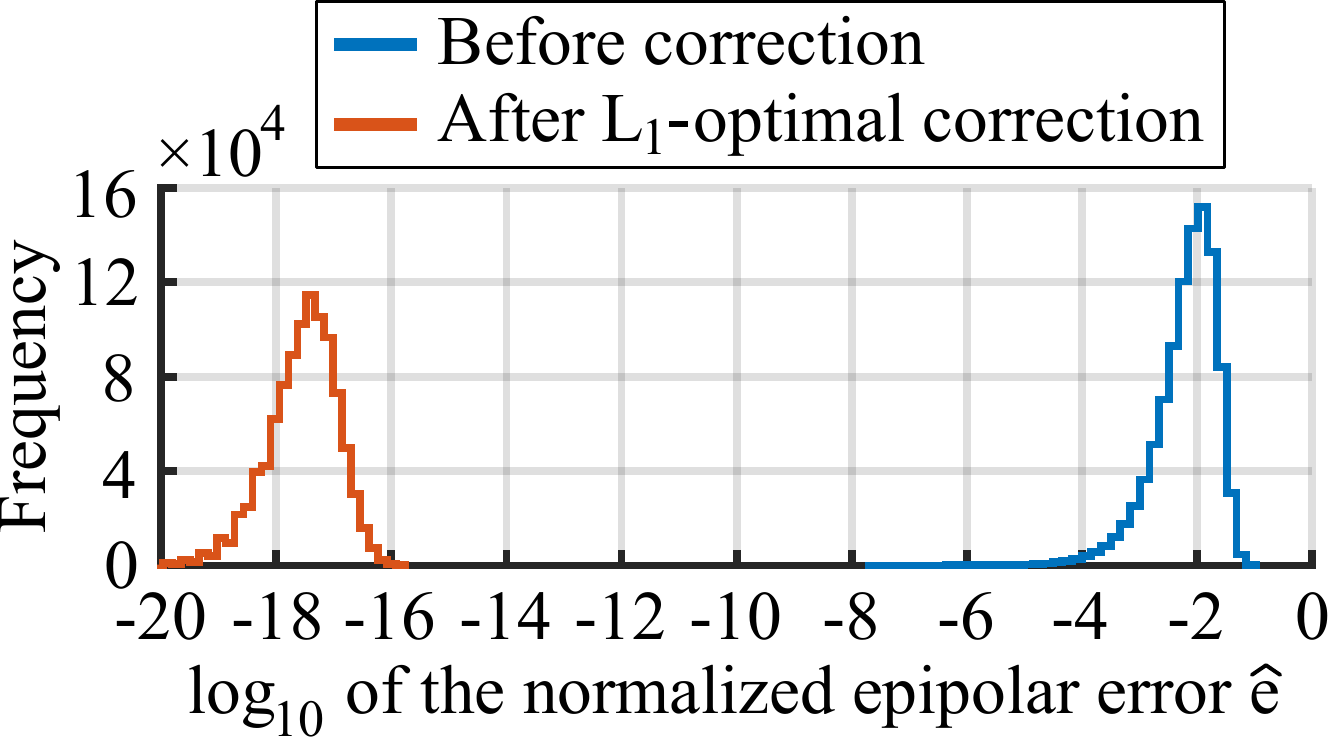}
\caption{Histograms of the normalized epipolar error $\widehat{e}\hspace{1pt}$ before and after the $L_1$-optimal correction based on angular errors \cite{closed_form_optimal_triangulation_based_angular_errors}. 
The corrected rays yield minuscule error, which implies that they now intersect (within the numerical accuracy).
 }
\label{fig:nee_histogram}
\end{figure}
\begin{figure}[t]
 \centering
 \includegraphics[width=0.34\textwidth]{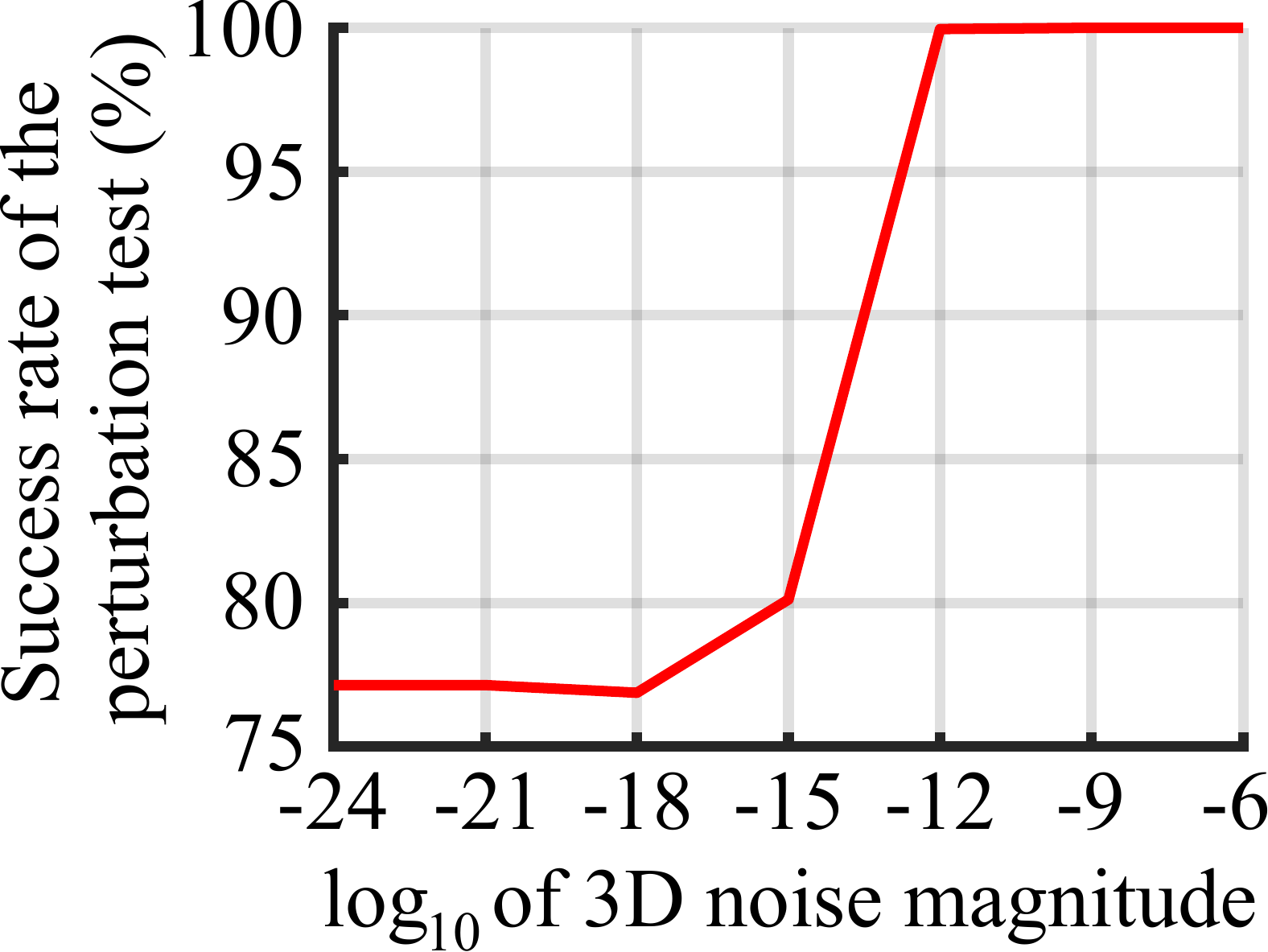}
\caption{The percentage of the simulation runs where the angular error $\theta^*_\mathrm{L1}$ obtained from the $L_1$-optimal method \cite{closed_form_optimal_triangulation_based_angular_errors} is smaller than that of the perturbed point.
In our experiment, $\theta^*_\mathrm{L1}$ is always smaller unless the triangulated point is perturbed by extremely small noise ($<10^{-12}$ unit) beyond the numerical accuracy.
 }
\label{fig:optimality}
\end{figure}

\section{Conclusion}
In this work, we presented several geometric interpretations of the normalized epipolar error $\widehat{e}$ defined in \eqref{eq:normalized_epipolar_error}.
Specifically, we revealed the direct relations between this error and the following quantities:
\begin{enumerate}\itemsep0em
    \item The volume of the tetrahedron where $\widehat{\mathbf{f}}_0$, $\mathbf{R}\widehat{\mathbf{f}}_0$ and $\widehat{\mathbf{t}}$ form the three edges meeting at one vertex (see Fig. \ref{fig:tetrahedron}{\color{red}a}).
    The relation is given by \eqref{eq:volume_interpretation}.
    \item The shortest distance between the two backprojected rays $\mathbf{l}_0=\mathbf{t}+s_0\mathbf{R}\widehat{\mathbf{f}}_0$ and $\mathbf{l}_1=s_1\widehat{\mathbf{f}}_1$ (see Fig. \ref{fig:tetrahedron}{\color{red}b}).
    The relation is given by \eqref{eq:distance_interpretation}.
    \item The dihedral angle between the two bounding epipolar planes, \ie, one plane containing $\mathbf{t}$ and $\mathbf{R}\mathbf{f}_0$ and the other containing $\mathbf{t}$ and $\mathbf{f}_1$ (see Fig. \ref{fig:skew_distance}{\color{red}a}). The relation is given by \eqref{eq:dihedral3}.
    \item The $L_1$-optimal angular reprojection error defined in \eqref{eq:L1_angle_definition}.
    The relation is given by \eqref{eq:L1_angle_interpretation}.
\end{enumerate}

\section*{Acknowledgement}
This work was partially supported by the Spanish government (project PGC2018- 096367-B-I00) and the Arag{\'{o}}n regional government (Grupo DGA-T45{\_}17R/FSE).

\newpage
\section*{Appendix}

\begin{figure}[t]
 \centering
 \includegraphics[width=0.33\textwidth]{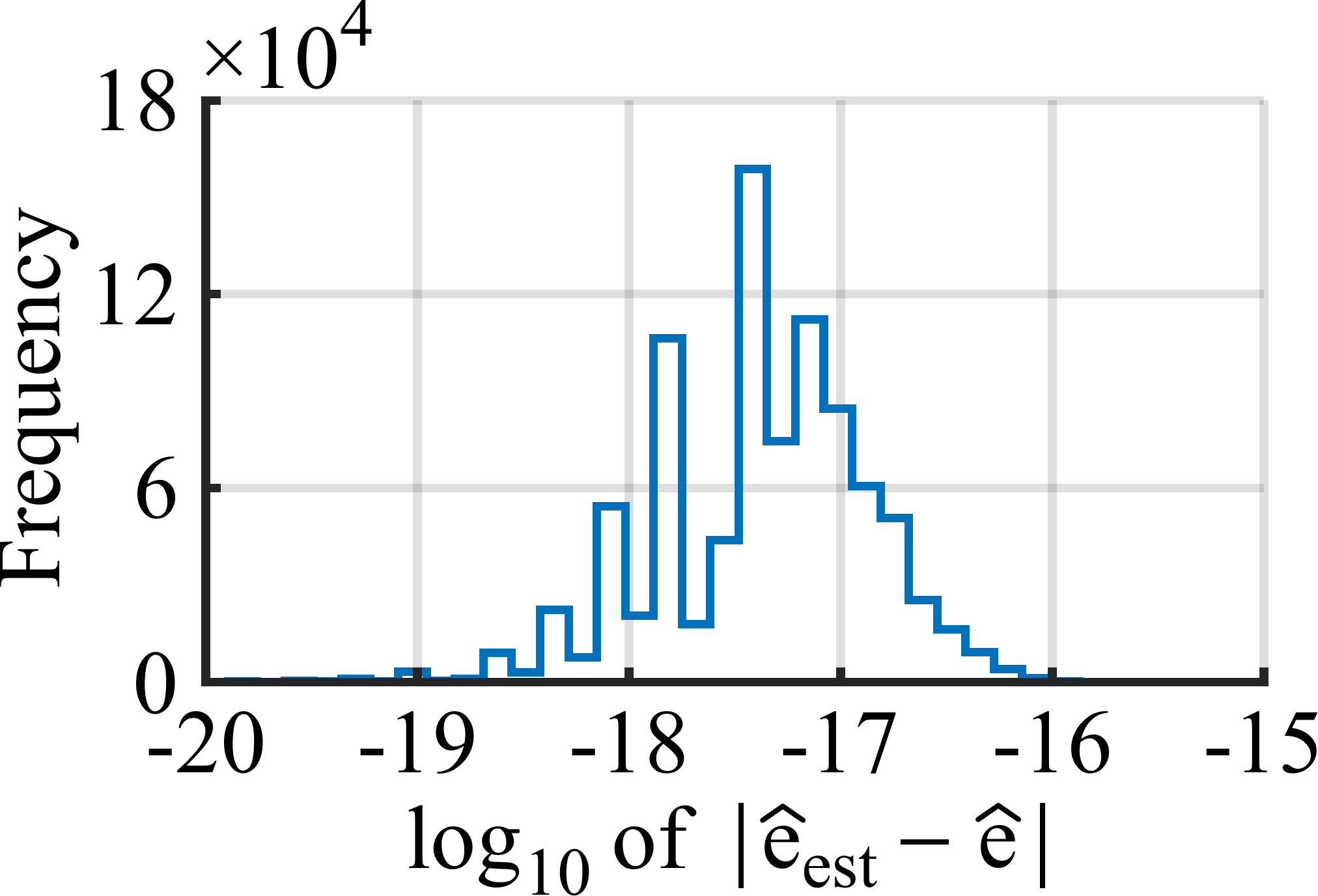}
\caption{Histogram of $|\widehat{e}_\text{est}-\widehat{e}\hspace{1pt}|$, where $\widehat{e}\hspace{1pt}$ and $\widehat{e}_\text{est}$ are the normalized epipolar errors computed using \eqref{eq:normalized_epipolar_error} and  \eqref{eq:L1_angle_interpretation} respectively. 
 }
\label{fig:nee_difference}
\end{figure}
\vspace{-0.5em}
\begin{figure}[t]
 \centering
 \includegraphics[width=0.33\textwidth]{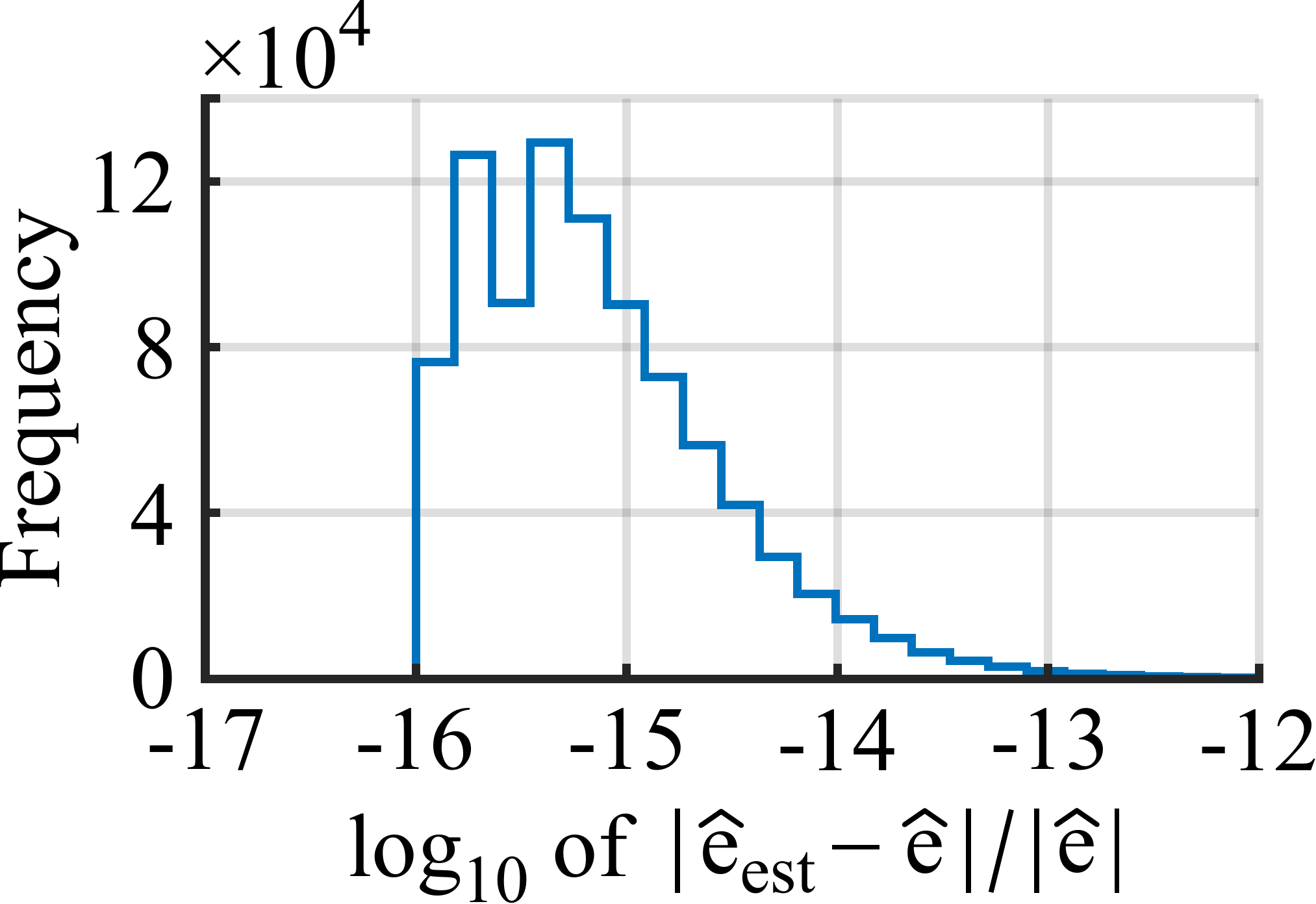}
\caption{Histogram of $|\widehat{e}_\text{est}-\widehat{e}\hspace{1pt}|/|\widehat{e}\hspace{1pt}|$, where $\widehat{e}\hspace{1pt}$ and $\widehat{e}_\text{est}$ are the normalized epipolar errors computed using \eqref{eq:normalized_epipolar_error} and  \eqref{eq:L1_angle_interpretation} respectively. 
 }
\label{fig:nee_difference2}
\end{figure}

The contributions of this work are the derivations of \eqref{eq:volume_interpretation}, \eqref{eq:distance_interpretation}, \eqref{eq:dihedral3} and \eqref{eq:L1_angle_interpretation}.
No approximation is made in the derivations, so strictly speaking, experiments are redundant as long as the mathematics are correct.
Having said that, we understand that some readers may have doubts about the derivations, and also, it is essential to verify the theoretical results whenever possible (as a sanity check).
In the case of \eqref{eq:volume_interpretation}, \eqref{eq:distance_interpretation} and \eqref{eq:dihedral3}, however, performing experiments is pointless because the only sensible method to compute the volume $V$, the distance $d$ and the angle $\alpha$ is to use the very same formulas used in the derivations.
For this reason, we only focus on the verification of \eqref{eq:L1_angle_interpretation} in this section.

In order to verify \eqref{eq:L1_angle_interpretation}, we compare the values of $\widehat{e}$ computed using \eqref{eq:normalized_epipolar_error} and \eqref{eq:L1_angle_interpretation}. 
This is done in the following steps:
\begin{enumerate}[leftmargin=*]\itemsep0em
    \item 
    In simulation, we create two cameras and one point.
    The two cameras are placed at position $\mathbf{c}_0$ and $\mathbf{c}_1$ where $\mathbf{c}_0$ is a random 3D vector of length 0.5 unit and $\mathbf{c}_1=-\mathbf{c}_0$.
    This ensures that $\lVert\mathbf{t}\rVert=\lVert\mathbf{c}_0-\mathbf{c}_1\rVert=1$ unit.
    The image size is set to $640\times480$ pixel and the focal length to $525$ pixel.
    We place the point at $[0, 0, D]^\top$ where $D$ follows the uniform distribution $\mathcal{U}(1, 10)$.
    Then, we orient the cameras randomly until the point is visible in both views.
    The image coordinates of the projected point are perturbed by Gaussian noise $\mathcal{N}(0, \sigma^2)$ with $\sigma=10$ pix. 
    
    \item
    We correct the backprojected rays using the $L_1$-optimal triangulation method described in \cite{closed_form_optimal_triangulation_based_angular_errors} and obtain the angular error $\theta^*_\mathrm{L1}$ using \eqref{eq:L1_angle_definition}.
    To check if this is locally optimal, we perturb the resulting 3D point by small random noise and see if we achieve smaller error.
    We set the noise magnitude to $10^m$ unit with $m\in\{-24, -21, \cdots, -6\}$, and for each magnitude, we perturb the point one hundred times independently.
    
    \item 
    We compute $\widehat{e}$ using \eqref{eq:L1_angle_interpretation} and compare it to $\widehat{e}$ from \eqref{eq:normalized_epipolar_error}.  
\end{enumerate}
We repeat this procedure $10^6$ times and aggregate the results.
All computations are done in Matlab.
Fig. \ref{fig:nee_histogram} shows the histograms of the normalized epipolar error $\widehat{e}$ computed using \eqref{eq:normalized_epipolar_error} before and after the $L_1$-optimal ray correction \cite{closed_form_optimal_triangulation_based_angular_errors}.
Comparing the two histograms, we see that the corrected rays do  intersect.
Fig. \ref{fig:optimality} presents the result of the perturbation test.
It shows that the angular error $\theta^*_\mathrm{L1}$ of the corrected rays is (locally) minimum within the numerical accuracy.
Plugging $\theta^*_\mathrm{L1}$ into \eqref{eq:L1_angle_interpretation}, we obtain an estimate of $\widehat{e}$, \ie, $\widehat{e}_\text{est}$.
In Fig. \ref{fig:nee_difference}, we plot the histogram of the absolute difference $|\widehat{e}_\text{est}-\widehat{e}\hspace{1pt}|$.
Notice that it is as small as the normalized epipolar error of intersecting rays (see Fig. \ref{fig:nee_histogram}).
Therefore, we can safely conclude that $\widehat{e}_\text{est}=\widehat{e}$ within the numerical accuracy.
In Fig. \ref{fig:nee_difference2}, we provide, for completeness, the histogram of the relative difference $|\widehat{e}_\text{est}-\widehat{e}\hspace{1pt}|/|\widehat{e}\hspace{1pt}|$.

\cleardoublepage
{\small
\balance
\bibliographystyle{ieee}
% \bibliography{egbib}

}

\end{document}